\documentclass[letterpaper]{article} 
\usepackage{aaai2026}  
\usepackage{times}  
\usepackage{helvet}  
\usepackage{courier}  
\usepackage[hyphens]{url}  
\usepackage{graphicx} 
\usepackage{natbib}  
\usepackage{caption} 
\usepackage{subcaption}
\usepackage{multirow}
\usepackage{booktabs} 
\usepackage{algorithm}
\usepackage{algorithmic}
\usepackage{amsmath} 
\DeclareUnicodeCharacter{2212}{\textminus}
\usepackage{newfloat}
\usepackage{listings}
\DeclareCaptionStyle{ruled}{labelfont=normalfont,labelsep=colon,strut=off} 
\floatstyle{ruled}
\newfloat{listing}{tb}{lst}{}
\floatname{listing}{Listing}
\title{NoReGeo: Non-Reasoning Geometry Benchmark}
\author {
     Irina Abdullaeva\textsuperscript{\rm 1,\rm 2},
     Anton Vasiliuk\textsuperscript{\rm 1},
     Elizaveta Goncharova\textsuperscript{\rm 1,\rm 3},
    Temurbek Rahmatullaev\textsuperscript{\rm 1,\rm 4}, \\
    Zagorulko Ivan\textsuperscript{\rm 5}, 
    Maxim Kurkin\textsuperscript{\rm 1, \rm 6}, 
    Andrey Kuznetsov\textsuperscript{\rm 1,\rm 2}
}
\affiliations {
    \textsuperscript{\rm 1}FusionBrain Lab, Russia\\
    \textsuperscript{\rm 2}Research Center of the Artificial Intelligence Institute, Innopolis University, Innopolis, Russia\\
    \textsuperscript{\rm 3}HSE University, Russia\\
    \textsuperscript{\rm 4}Lomonosov Moscow State University, Russia\\
    \textsuperscript{\rm 5}Central University, Russia\\
    \textsuperscript{\rm 6}Applied AI Institute, Moscow, Russia\\
    abdullaeva@fusionbrainlab.com, goncharova@fusionbrainlab.com, kuznetsov@fusionbrainlab.com
}

\usepackage{bibentry}

\usepackage{tcolorbox}
\begin{document}

\maketitle

\begin{abstract}
We present NoReGeo, a novel benchmark designed to evaluate the intrinsic geometric understanding of large language models (LLMs) without relying on reasoning or algebraic computation. Unlike existing benchmarks that primarily assess models' proficiency in reasoning-based geometry-where solutions are derived using algebraic methods-NoReGeo focuses on evaluating whether LLMs can inherently encode spatial relationships and recognize geometric properties directly. Our benchmark comprises 2,500 trivial geometric problems spanning 25 categories, each carefully crafted to be solvable purely through native geometric understanding, assuming known object locations. We assess a range of state-of-the-art models on NoReGeo, including frontier models like GPT-4, observing that even the most advanced systems achieve an overall maximum of 65\% accuracy in binary classification tasks. Further, our ablation experiments demonstrate that such geometric understanding does not emerge through fine-tuning alone, indicating that effective training for geometric comprehension requires a specialized approach from the outset. Our findings highlight a significant gap in current LLMs' ability to natively grasp geometric concepts, providing a foundation for future research toward models with true geometric cognition. 
\end{abstract}

\begin{links}
    \link{Code}{https://github.com/FusionBrainLab/NoReGeo}
\end{links}

\begin{figure*}[ht]
  \centering
  \includegraphics[width=0.9\textwidth]{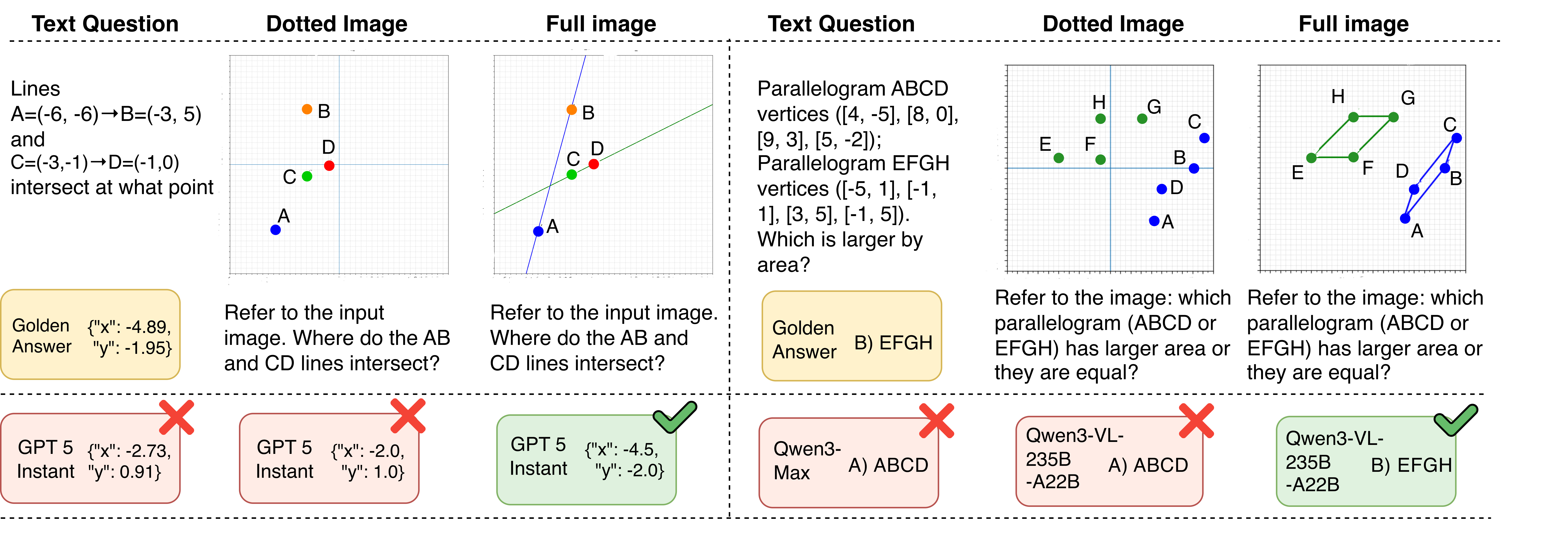}
  \caption{Evaluation samples from NoReGeo benchmark. Each problem is shown in three formats -- (a) text‑only, (b) text with dotted-image (points only), and (c) text with full-image (points plus connecting lines) -- together with the golden answer (yellow) and the model’s prediction.}
  \label{fig:scheme}
\end{figure*}

\section{Introduction}

Although modern LLMs excel at symbolic reasoning, they still treat even the simplest spatial relations as formal reasoning problems, producing multi-step chains of thought instead of relying on an intuitive sense of geometry. Ask whether two line segments intersect, and many models unfold a miniature proof rather than making a direct geometric judgment. This gap between symbolic reasoning and intuitive spatial understanding becomes especially limiting in time-critical settings — CAD engines updating thousands of vertices per frame, robots refining grasp trajectories in milliseconds, or geospatial systems performing rapid visibility checks. In such environments, approximate geometric intuition is far more valuable than multi-line chains of thought. Yet current models lack native geometric understanding and must simulate reasoning even for the simplest relations.

Existing geometry benchmarks primarily emphasize complex proofs or multi-step algebraic reasoning \cite{geometry3k,geomverse}, conflating two skills: (i) identifying relevant spatial facts and (ii) executing symbolic derivations. To isolate the first skill, we introduce \textbf{NoReGeo}, a benchmark of 2,500 trivially solvable geometry problems across 25 categories. Each item can be answered directly from point locations, without auxiliary constructions, theorems, or lengthy CoT. Every problem appears in both text-only form and a paired diagram, enabling controlled comparisons between text-based LLMs and multimodal vision-language models (VLMs).

We evaluate more than 45 frontier and open-source models under both modalities. Even the strongest open-source VLM achieves only $\sim\!55\%$ accuracy (Phi3.5-Vision), and the best proprietary model reaches $\sim\!65\%$ — well below human performance of $\sim\!74.5\%$ on the same multiple-choice tasks. These results expose a substantial gap in basic spatial intuition. Further analysis shows that fine-tuning alone does not confer geometric competence, whereas a simple linear probe on a frozen vision encoder solves the tasks almost perfectly — suggesting that geometric features are present in embeddings but are not accessed by current LLM architectures or training regimes.

Our contributions are as follows:
\begin{enumerate}
    \item We motivate and formalize the concept of \emph{native geometric understanding} as a core capability required for spatially intensive applications.
    \item We introduce \textbf{NoReGeo}, the first benchmark explicitly designed to test this ability without chain-of-thought or algebraic computation, comprising 2\,500 items across 25 categories in both text and image formats.
    \item We provide a comprehensive evaluation of over 45 state-of-the-art LLMs and VLMs, showing that all fall short of human-level performance on elementary geometric tasks.
    \item Through fine-tuning and linear-probing studies, we demonstrate that geometric knowledge exists latently in vision encoders but fails to naturally emerge in current LLM training paradigms.
\end{enumerate}

Overall, our findings point to an urgent open problem: \textbf{\emph{bridging the gap between symbolic reasoning and true geometric cognition in future foundation models.}}


\section{Related Work}\label{sec:related}

\paragraph{Spatial reasoning in vision–language models.}
Vision–language systems must ground textual concepts such as \emph{left of} or \emph{bigger than} in a spatial frame of reference.  
Synthetic VQA benchmarks like \textsc{CLEVR} isolate this ability by rendering scenes of coloured shapes and asking queries that combine Boolean logic with coarse spatial predicates \cite{clevr}.  
Subsequent datasets move toward natural imagery: \textsc{SpatialSense} adversarially mines object pairs (e.g.\ \emph{person–bench}) to evaluate relative positioning \cite{yang2019spatialsense}, while \textsc{3DSRBench} embeds similar relations in RGB-D scans of indoor environments \cite{ma20253dsrbench}.  
Interactive navigation tasks extend spatial grounding to embodied settings: BabyAI and ALFRED require locating, manipulating, and placing objects in simulated rooms \cite{babyai,alfred}, and Room-to-Room (R2R) tests natural-language navigation in realistic 3-D reconstructions \cite{r2r}.  
Together, these datasets show that VLMs handle \emph{qualitative} spatial relations reasonably well, yet they provide little evidence that models encode the \textbf{fine-grained geometric attributes} — distances, angles, midpoints, coordinate relations — needed for the intuitive geometric understanding.

\paragraph{Geometry word-problem benchmarks.}
A parallel line of work examines whether models can solve textbook geometry questions that pair a diagram with natural language.  
Benchmarks such as \textsc{Geometry3K} \cite{geometry3k}, \textsc{PGPS-9K} \cite{pgps9k}, MathVista \cite{lu2024mathvista}, MathVerse \cite{zhang2024mathverse}, and \textsc{GeomVerse} \cite{geomverse} collect thousands of K–12 geometry problems, while \textsc{GeoEval} adds a calibrated difficulty ladder and reports that even math-tuned LLMs plateau at \(\approx55\%\) accuracy on ``regular'' items and \(<10\%\) on Olympiad-level ones \cite{geoeval}.  
Solving these tasks typically requires \emph{multi-step reasoning}: interpreting diagrams, identifying theorems, and chaining symbolic deductions.  
As a result, existing evaluations emphasize algebraic manipulation and proof-style solution pipelines — often supported by chain-of-thought prompting, external tools, or full symbolic proof synthesis (e.g.\ AlphaGeometry \cite{alphageometry}) — making it difficult to determine whether models possess any \emph{native geometric perception} independent of reasoning.

\paragraph{Skill-specific probes.}
To obtain a more targeted diagnostic signal, several recent efforts isolate individual geometric skills.  
\textsc{PlanQA} presents floor-plan layouts as JSON and asks questions about visibility or shortest paths \cite{planqa}.  
\textsc{GeomRel} focuses on detecting equal segments and angles before any numeric computation \cite{geomrel}, and \textsc{GeoGramBench} converts Asymptote-style code into natural-language questions of varying abstraction \cite{geogrambench}.  
These benchmarks remove some confounds but still require models to parse code-like input formats or engage in subtle inference — leaving open the question of whether models can make \emph{direct, perception-level} geometric judgments.

The proposed cross-modal \textsc{NoReGeo} benchmark targets \textbf{single-step, school-level geometry questions} (midpoints, area comparisons, collinearity, symmetry tests, and similar micro-skills) that can be answered instantaneously by anyone with basic geometric intuition with no auxiliary constructions, theorem recall, or multi-hop reasoning.  
By providing both text-only and vision‑augmented variants under a shared output format, \textsc{NoReGeo} enables controlled comparisons between LLMs and VLMs and reveals whether either modality supplies genuine geometric understanding.  
Because the tasks are \emph{synthetic yet curriculum‑aligned}, they are unlikely to appear verbatim in pre-training corpora, ensuring that NoReGeo is a genuine test of latent geometric understanding rather than memorization. 

Thus, our benchmark complements prior work by stripping away reasoning scaffolds and focusing on the bedrock geometric knowledge that more complex systems implicitly assume.

\section{Benchmark Motivation and Scope}

The NoReGeo is designed as a cross-modal geometry-based benchmark oriented to probe the \textbf{foundational geometric competence} of modern LLMs and VLMs. The benchmark consists of short prompt–answer pairs in elementary geometry. Each problem is posed as a one‑shot query: the model receives a single prompt and must return an answer immediately, without any chain‑of‑thought or intermediate steps. The benchmark provides two prompt modalities — text‑only questions and their corresponding image‑based versions (where the images are presented in the so-called dotted and full versions). The example of samples from the NoReGeo can be found in Figure \ref{fig:scheme}.

Each problem has a ground-truth answer that is either a numeric value (for quantitative questions, typically an integer or simple fraction; e.g. 90) or a categorical label (for qualitative classification questions; e.g. "acute" to describe an angle type). There are no elaborate proofs or explanations required -- the output is a single final answer. The evaluation metric is straightforward accuracy for multiple-choice questions and soft accuracy (within the [-0.5, 0.5] interval) for numeric ones.

\subsection{High‑Level Taxonomy of Tasks}\label{sec:taxonomy}

During benchmark creation, we followed the typical secondary school geometry curriculum, our benchmark effectively covers content taught from roughly higher school \cite{CCSS2010,NCTM2000}. We have included some foundational topics that are introduced in middle school (for example, basic angle facts or simple constructions from early secondary years) as well as the full suite of high-school geometry topics (Euclidean proofs, circle theorems, introductory trigonometry, etc.) The detailed taxonomy of NoReGeo is given in Table \ref{tab:bench_tasks}.

The benchmark tasks are categorized into three types: \emph{Classification}, \emph{Numeric}, and \emph{Unstable}. Classification tasks (C‑) involve multiple-choice questions, such as identifying polygon areas or symmetry. Numeric tasks (N‑) require numeric values like coordinates or lengths, while Unstable tasks (U‑) involve binary decisions that can change with minimal input variation.

\subsection{Dataset Construction}

Building on the taxonomy described above, we developed a concise pipeline for synthetic data generation. Each benchmark item is provided either in text-only form or as a multimodal (vision–text) variant combining text with an image.

All items in the dataset follow these design rules:

\begin{itemize}
    \item Every vertex uses integer coordinates in the range $[-20,20]$. When an image is present, the points lie on a Cartesian grid.
    \item In text-only prompts, points are denoted by uppercase letters with their coordinates, e.g.\ $A=(2,1)$.
\end{itemize}

The multimodal format has two variants:  
(i) an image showing only labeled points (without coordinates labels), with edges implied by the text; or  
(ii) an image of the complete figure, where the text provides only the question (and any answer options) without listing coordinates. The sample questions are give below.

\begin{flushleft}
\textit{Text-only.} \quad Lines $A=(2,1)\!\rightarrow\!B=(3,0)$ and $C=(-8,-1)\!\rightarrow\!D=(9,0)$ intersect at what point?\\[2pt]
\textit{Multimodal.} \quad Refer to the input image. Where do lines $AB$ and $CD$ intersect?
\end{flushleft}

We split the questions into \textit{vision-only} and \textit{text-only} tasks to minimize the influence of the textual prompt on the vision-based tasks, ensuring that coordinates must be read solely from the image that is a significant challenge in modern multimodal benchmarks \cite{chen2024rightwayevaluatinglarge}.

\begin{table*}[h]
\centering
{\fontsize{9pt}{11pt}\selectfont
\begin{tabular}{@{}lp{3cm}p{2.4cm}lp{0.8cm}p{6.3cm}@{}}
\toprule
\textbf{Type} & \textbf{Category} & \textbf{Task} & \textbf{ID} & \textbf{Type} & \textbf{Sample Question} \\
\midrule
\multirow{4}{*}{Class.}
    & Area comparison & parallelogram\_size & C-ACM-PST & MC & Parallelograms ABCD and EFGH (vertices given). Which has larger area? \\
    &                 & triangle\_size       & C-ACM-TST & MC & Triangles ABC and DEF. Which has larger area? \\
    & Basic coordinate tasks & collinearity & C-BCT-CT & MC & Are points A, B, and C collinear? \\
    & Symmetry        & shape\_symmetry     & C-SYM-SST & MC & Polygon (vertices given): symmetric about Y-axis? \\
\midrule
\multirow{12}{*}{Numeric}
    & Basic coordinate tasks & midpoint         & N-BCT-MT & Coord & Find midpoint of segment AB. \\
    &                        & parallelogram\_area    & N-BCT-PAT & Num & Parallelogram ABCD: find area. \\
    &                        & parallelogram\_perim & N-BCT-PPT & Num & Parallelogram ABCD: find perimeter. \\
    &                        & triangle\_area     & N-BCT-TAT & Num & Triangle ABC: find area. \\
    &                        & triangle\_perim & N-BCT-TPT & Num & Triangle ABC: find perimeter. \\
    \cmidrule(l){2-5}
    & Elementary calc.      & intersection        & N-ECL-IT  & Coord & Lines AB and CD: find intersection point. \\
    &                         & segment\_length     & N-ECL-SLT & Num & Segment AB: find length. \\
    \cmidrule(l){2-5}
    & Geometric transform. & reflection       & N-GTR-RT  & Coord & Reflect point P across X-axis. Find coordinates. \\
    &                            & rotation\_point  & N-GTR-RPT & Coord & Rotate point P 90° about origin. \\
    \cmidrule(l){2-5}
    & Remarkable triangle lines & bisector         & N-RLT-BT  & Coord & Triangle ABC: find B-angle bisector intersection. \\
    \cmidrule(l){2-5}
    & Simple circle properties
        & inner\_circle\_center & N-SCP-ICCT & Coord & Triangle ABC: find incircle center. \\
    &        & inner\_circle\_radius & N-SCP-ICRT & Num & Triangle ABC: find incircle radius. \\
    &        & outer\_circle\_center & N-SCP-OCCT & Coord & Triangle ABC: find circumcenter. \\
    &        & triangle\_type        & N-SCP-TTT  & Num & Triangle ABC: find triangle type. \\
\midrule
\multirow{6}{*}{Unstable}
    & Circle properties         & circle\_diameter     & U-CPR-CDT & MC & Circle with center O, radius = 3; is AB a diameter? \\
    \cmidrule(l){2-5}
    & Parallelism & parallel\_lines & U-PAP-PLT & MC & Are lines AB and CD parallel? \\
    & and Perpendicularity & perpendicular & U-PAP-PT & MC & Are segments AB and CD perpendicular? \\
    &                       & right\_angle & U-PAP-RAT & MC & Is angle ABC a right angle? \\
    \cmidrule(l){2-5}
    & Remarkable triangle lines & special\_lines & U-RLT-SLT & MC & Triangle ABC with segment from B to D. Identify the segment. \\
    \cmidrule(l){2-5}
    & Simple circle properties & semicircle\_triangle & U-SCP-STT & MC & Is triangle ABC inscribed in a semicircle? \\
    \cmidrule(l){2-5}
    & Symmetry & symmetry & U-SYM-ST & MC & Are points P and Q symmetric about line y=x? \\
\bottomrule
\end{tabular}%
}
\caption{Overview of the NoReGeo benchmark tasks, organized by question type, geometric category, and specific task. Each task is identified by a structured ID code consisting of: (i) a single-letter type prefix (C: Classification, N: Numeric, U: Unstable), (ii) a three-letter category code, and (iii) a short task name. The benchmark covers a broad range of geometric reasoning tasks including area and perimeter comparison, coordinate calculations, symmetry, triangle centers, and more.}
\label{tab:bench_tasks}
\end{table*}

As shown in Table \ref{tab:bench_tasks}, the benchmark maintains a slight emphasis on basic coordinate geometry over symmetry tasks. Notably, there is a significant majority of classification tasks compared to numerical tasks. This design choice prioritizes the evaluation of a model's ability to recognize geometric properties and apply definitions over pure computation. All problems are solvable through the direct application of fundamental formulas and definitions, making the benchmark a robust tool for assessing core geometric understanding in AI models across both text and vision modalities.

\section{Experiments}

In this section, we evaluate whether LLMs and VLMs can natively perceive geometric structures and relationships across varying text-to-visual information ratios using our NoReGeo benchmark.

\begin{table*}[]
\centering
{\fontsize{9pt}{11pt}\selectfont
\begin{tabular}{cccccccccccccc}
 & \multicolumn{3}{c|}{\textbf{Classification}} & \multicolumn{5}{c|}{\textbf{Numeric}} & \multicolumn{5}{c}{\textbf{Unstable}} \\ \cline{2-14} 
 & ACM & BCT & \multicolumn{1}{c|}{SYM} & BCT & ECL & GTR & RLT & \multicolumn{1}{c|}{SCP} & CPR & PAP & RLT & SCP & SYM \\ \hline
\multicolumn{14}{c}{\textbf{Text}} \\ \hline
Qwen2.5-3B-In. & \textbf{69.2} & 44.0 & \multicolumn{1}{c|}{54.0} & 3.9 & \textbf{32.3} & 0.0 & \textit{98.5} & \multicolumn{1}{c|}{\textit{32.2}} & \textit{53.0} & 47.8 & 34.5 & 0.0 & 46.5 \\
Qwen2.5-7B-In. & 24.5 & 52.0 & \multicolumn{1}{c|}{57.0} & 5.6 & 12.7 & \underline{\textbf{92.0}} & \underline{\textbf{99.0}} & \multicolumn{1}{c|}{21.2} & 6.0 & 21.3 & 22.0 & 7.0 & 47.0 \\
Qwen3-4B & 56.5 & 61.0 & \multicolumn{1}{c|}{52.0} & 5.0 & \textit{17.0} & 0.0 & 40.0 & \multicolumn{1}{c|}{\textbf{36.2}} & 52.0 & 52.7 & 31.0 & 49.0 & \textit{89.0} \\
Qwen3-8B & 58.0 & \textit{73.0} & \multicolumn{1}{c|}{67.0} & \textbf{6.8} & 10.7 & 0.0 & 4.0 & \multicolumn{1}{c|}{25.8} & 52.0 & \textit{53.0} & \textbf{47.0} & \textit{64.0} & \textbf{91.0} \\
\begin{tabular}[c]{@{}c@{}}Mistral-Small-In.\end{tabular} & 57.5 & 56.0 & \multicolumn{1}{c|}{59.0} & 3.8 & 12.0 & 0.0 & 3.0 & \multicolumn{1}{c|}{14.5} & 52.0 & 45.3 & 32.0 & 50.0 & 79.0 \\
\begin{tabular}[c]{@{}c@{}}LLaMa-3.1 8B-In.\end{tabular} & 64.5 & 51.0 & \multicolumn{1}{c|}{\textit{96.0}} & 1.6 & 9.3 & 0.0 & 18.0 & \multicolumn{1}{c|}{15.2} & \textit{53.0} & 52.0 & 40.0 & \textbf{67.0} & 68.0 \\
\begin{tabular}[c]{@{}c@{}}LLaMa-3.1 70B-In.\end{tabular} & \textit{67.0} & \textbf{78.0} & \multicolumn{1}{c|}{\textbf{97.0}} & \textit{6.4} & 16.7 & 0.0 & 0.0 & \multicolumn{1}{c|}{17.2} & \textbf{97.0} & \textbf{53.7} & \textit{43.0} & 52.0 & 84.0 \\ \hline
\multicolumn{14}{c}{\textbf{Text with dotted images}} \\ \hline
Qwen2-VL-7B-In. & 38.5 & 54.0 & \multicolumn{1}{c|}{52.0} & 9.4 & 35.7 & \textit{50.5} & \textit{50.0} & \multicolumn{1}{c|}{\textit{57.2}} & 52.0 & \textit{50.0} & 76.0 & 49.0 & 49.0 \\
Qwen2.5-VL-7B-In. & 40.5 & 51.0 & \multicolumn{1}{c|}{47.0} & 3.4 & 12.3 & 0.0 & 1.0 & \multicolumn{1}{c|}{31.5} & 52.0 & 48.3 & 92.0 & 43.0 & 50.0 \\
InternVL2.5-8B & 34.0 & 59.0 & \multicolumn{1}{c|}{52.0} & \textit{18.6} & \underline{\textbf{60.7}} & 8.5 & 28.0 & \multicolumn{1}{c|}{63.0} & 52.0 & 43.7 & 37.0 & 58.0 & 46.0 \\
InternVL3-8B & \textit{41.0} & 53.0 & \multicolumn{1}{c|}{52.0} & 2.8 & 5.3 & 0.0 & 17.0 & \multicolumn{1}{c|}{39.2} & 52.0 & 55.7 & \underline{\textbf{99.0}} & 51.0 & 54.0 \\
\begin{tabular}[c]{@{}c@{}}LLaVA-Mini (LLaMA-8B)\end{tabular} & 20.5 & \textit{62.0} & \multicolumn{1}{c|}{59.0} & 20.8 & \textit{56.7} & \textbf{75.0} & \textbf{90.0} & \multicolumn{1}{c|}{43.5} & 53.0 & 47.0 & 29.0 & 52.0 & 39.0 \\
MiniCPM-o-2.6 & 40.0 & 44.0 & \multicolumn{1}{c|}{\textit{54.0}} & 7.8 & 17.0 & 0.0 & 2.0 & \multicolumn{1}{c|}{48.5} & 44.0 & \textit{50.0} & 42.0 & 49.0 & 53.0 \\
\begin{tabular}[c]{@{}c@{}}Phi-3.5-Vis.-In.\end{tabular} & 41.5 & \textbf{73.0} & \multicolumn{1}{c|}{52.0} & 15.4 & 32.3 & 1.0 & 35.0 & \multicolumn{1}{c|}{46.5} & 52.0 & 49.3 & \textit{93.0} & \textit{55.0} & 50.0 \\ 
\hline
Human eval & \underline{\textbf{81.5}} & 70.0 & \multicolumn{1}{c|}{\textbf{72.0}} & \textbf{63.0} & 0.0 & 5.0 & 50.0 & \multicolumn{1}{c|}{\underline{\textbf{78.5}}} & \textbf{88.0} & \underline{\textbf{92.0}} & 89.0 & \textbf{81.0} & \underline{\textbf{94.0}} \\
\hline
\multicolumn{14}{c}{\textbf{Text with full images}} \\ \hline
Qwen2-VL-7B-In. & \textit{66.0} & 88.0 & \multicolumn{1}{c|}{90.0} & \underline{\textbf{81.8}} & \textbf{40.0} & \textit{50.0} & \textit{16.0} & \multicolumn{1}{c|}{36.0} & \underline{\textbf{100.0}} & 55.7 & 71.0 & \textit{99.0} & \textbf{86.0} \\
Qwen2.5-VL-7B-In. & \textbf{66.5} & \underline{\textbf{100.0}} & \multicolumn{1}{c|}{\textit{99.0}} & \textit{79.0} & 33.7 & 0.0 & 0.0 & \multicolumn{1}{c|}{26.8} & \underline{\textbf{100.0}} & \textit{63.0} & \textbf{98.0} & \textit{99.0} & \textit{62.0} \\
InternVL2.5-8B & 53.5 & 96.0 & \multicolumn{1}{c|}{67.0} & 75.6 & 32.7 & 1.5 & 10.0 & \multicolumn{1}{c|}{\textbf{44.0}} & 73.0 & 50.7 & 41.0 & 88.0 & 60.0 \\
\begin{tabular}[c]{@{}c@{}}LLaVA-Mini (LLaMA-8B)\end{tabular} & 24.5 & 52.0 & \multicolumn{1}{c|}{57.0} & 5.6 & 12.7 & \underline{\textbf{92.0}} & \underline{\textbf{99.0}} & \multicolumn{1}{c|}{21.2} & 6.0 & 21.3 & 22.0 & 7.0 & 47.0 \\
MiniCPM-o-2.6 & 52.5 & 88.0 & \multicolumn{1}{c|}{86.0} & 77.0 & 31.0 & 0.0 & 0.0 & \multicolumn{1}{c|}{17.5} & \textit{97.0} & \textbf{64.0} & 63.0 & 96.0 & 57.0 \\
\begin{tabular}[c]{@{}c@{}}Phi-3.5-Vis.-In.\end{tabular} & 50.0 & \textit{99.0} & \multicolumn{1}{c|}{\underline{\textbf{100.0}}} & 57.6 & \textit{39.3} & 0.0 & 13.0 & \multicolumn{1}{c|}{\textit{41.8}} & 65.0 & 52.3 & \textit{94.0} & \underline{\textbf{100.0}} & \textit{62.0} \\ \hline
\end{tabular}
}
\caption{Accuracy (\%) of selected models on each benchmark task category across three setups: text-only, text with dotted and full images. \textbf{Bold} indicates best per setup, \textit{italic} -- second-best, and \underline{underlined} shows overall best across all setups.}
\label{tab:main-results}
\end{table*}

\begin{table*}[h]
    \centering
    \begin{tabular}{lccc}
    \toprule
    \textbf{Task category} & \textbf{Text-only} & \textbf{Text with dot images} & \textbf{Text with full images} \\
    \midrule
    Area comparison (ACM) & 55.1 ± 19.1 & 33.6 ± 10.2 & 50.6 ± 23.5 \\
    Basic coordinate tasks (BCT) & 32.4 ± 39.5 & 19.2 ± 25.9 & 53.4 ± 42.8 \\
    Circle properties (CPR) & 59.4 ± 22.3 & 51.5 ± 4.6 & 71.0 ± 32.3 \\
    Elementary calculations (ECL) & 26.1 ± 36.6 & 38.3 ± 35.9 & 33.7 ± 37.9 \\
    Geometric transformations (GTR) & 12.1 ± 31.2 & 27.9 ± 36.6 & 18.6 ± 34.6 \\
    Parallelism and Perpendicularity (PAR) & 51.6 ± 15.6 & 48.8 ± 9.1 & 49.7 ± 20.1 \\
    Remarkable lines of a triangle (RLT) & 32.8 ± 30.9 & 57.3 ± 31.4 & 42.8 ± 37.0 \\
    Simple circle properties (SCP) & 35.2 ± 34.4 & 47.6 ± 28.2 & 39.4 ± 30.9 \\
    Symmetry (SYM) & 65.4 ± 19.6 & 50.0 ± 8.3 & 68.9 ± 27.5 \\
    \bottomrule
    \end{tabular}
    \caption{Average accuracy (\%) and standard deviation across models for each task category under different evaluation setups.}
    \label{tab:category_agg_results}
\end{table*}

\subsection{Experimental Setup}

\textbf{Models and Implementation.} We selected a broad range of state-of-the-art LLMs and VLMs, including both proprietary and open-source models, to capture representative examples of current capabilities. Our model selection also allows us to analyze trends across model generations and explore the effect of model scale on geometric and multimodal reasoning. In total, we evaluated over 45 models. Proprietary API-based models include GPT-4.1, GPT-4.1-Mini, and GPT-4.1-Nano. Open-source models span the Qwen series (Qwen2, Qwen2.5 \cite{team2024qwen2} and Qwen3 \cite{yang2025qwen3}, along with multimodal variants Qwen2-VL \cite{wang2024qwen2} and Qwen2.5-VL \cite{bai2025qwen2}), the LLaMA-3.1 series \cite{grattafiori2024llama}, and Mistral models \cite{jiang2023mistral7b} (including versions with math-specific pretraining), among others. Additionally, we evaluated math-specific VLMs, including the G-LLaVA \cite{gao2023g}, URSA \cite{luo2025ursa}, Math-LLaVA \cite{shi2024math}, and Multimath-7B-LLaVA-v1.5 \cite{peng2024multimath}, to contrast general-purpose and specialized models.

To ensure consistency across models, we standardized generation settings: fixed random seed, temperature set to 0.6, and a maximum output length of 2048 tokens. For instruction-tuned models, we used their native chat templates and applied a unified system prompt.

\noindent \textbf{Evaluation Scheme.} We focus on each model’s direct-answer capability - its ability to solve tasks without producing intermediate reasoning steps. To enforce structured output and prevent unsolicited reasoning, we applied a structured generation approach. Each task specifies a JSON-formatted response template based on the expected answer type (e.g., multiple choice, numeric value, or coordinate point). This structure is communicated to the model through a structure prompt appended to each question. 

We implemented this setup using the Outlines \cite{willard2023efficient} and xgrammar \cite{dong2024xgrammar} libraries, which convert expected JSON structures into regular expressions. These are compiled into finite state machines that bias model generation by modifying logits. For efficient model serving, we used the VLLM library \cite{kwon2023VLLM}.

\noindent \textbf{Evaluation Metrics \& Policy.} We evaluated the capabilities of LLMs and VLMs by comparing generated and reference answers using the accuracy metric. For multiple-choice tasks, answers were considered to match if they were an exact match for one of the generated answer options. For numerical and coordinate answers, we defined a tolerable error interval of 0.5, identical to the grid step in visualizations of geometric problems. Numerical answers and point coordinates were considered correct if they met the following criteria: a) they were valid numbers; b) they fell within the interval [reference answer - 0.5, reference answer + 0.5]. Regarding point coordinates, both coordinates of the answer point also had to be correct according to the numerical answer criteria mentioned above. To gain a more detailed understanding of error magnitude in cases involving numerical answers, we calculated regression metrics (mean squared error, MSE) for tasks where the answer was either a number or a point.

If the answer did not match the required format — if it was not valid JSON, included an additional reasoning trail, or lacked the correct answer fields in JSON — we marked it as incorrect and awarded no credit, even if the answer was mathematically correct. This strict policy isolates a model's geometric competence from its propensity to reveal private reasoning.

\paragraph{Human Evaluation}
For the dotted image format, we also conducted a human evaluation to assess how accurately humans can solve these tasks. The human baseline was obtained using the Toloka platform. Annotators completed training, examination, and control tasks, with 10 tasks per page and 10 minutes allowed per page. Each task was answered by three crowd workers, and majority voting was used to aggregate responses. Participants were instructed not to use external resources; the only aid was the task’s dotted-format overlay. The average annotator age was 39 years, with compensation of approximately \$1 per page.

\subsection{Main results}
\label{sec:reference_examples}

\textbf{General performance.} There is a significant disparity in the extent to which different models comprehend geometry and leverage cross-modal relations. Table \ref{tab:main-results} shows the average quality of a few representative LLMs and VLMs across task categories. Table \ref{tab:category_agg_results} shows the average quality of problem solving across all evaluated models within task categories, taking into account the standard deviation. Based on these results, we draw several key conclusions.

\paragraph{Full visual context significantly boosts VLM performance.}

\begin{figure}[t!] 
\centering
\includegraphics[width=0.98\columnwidth]{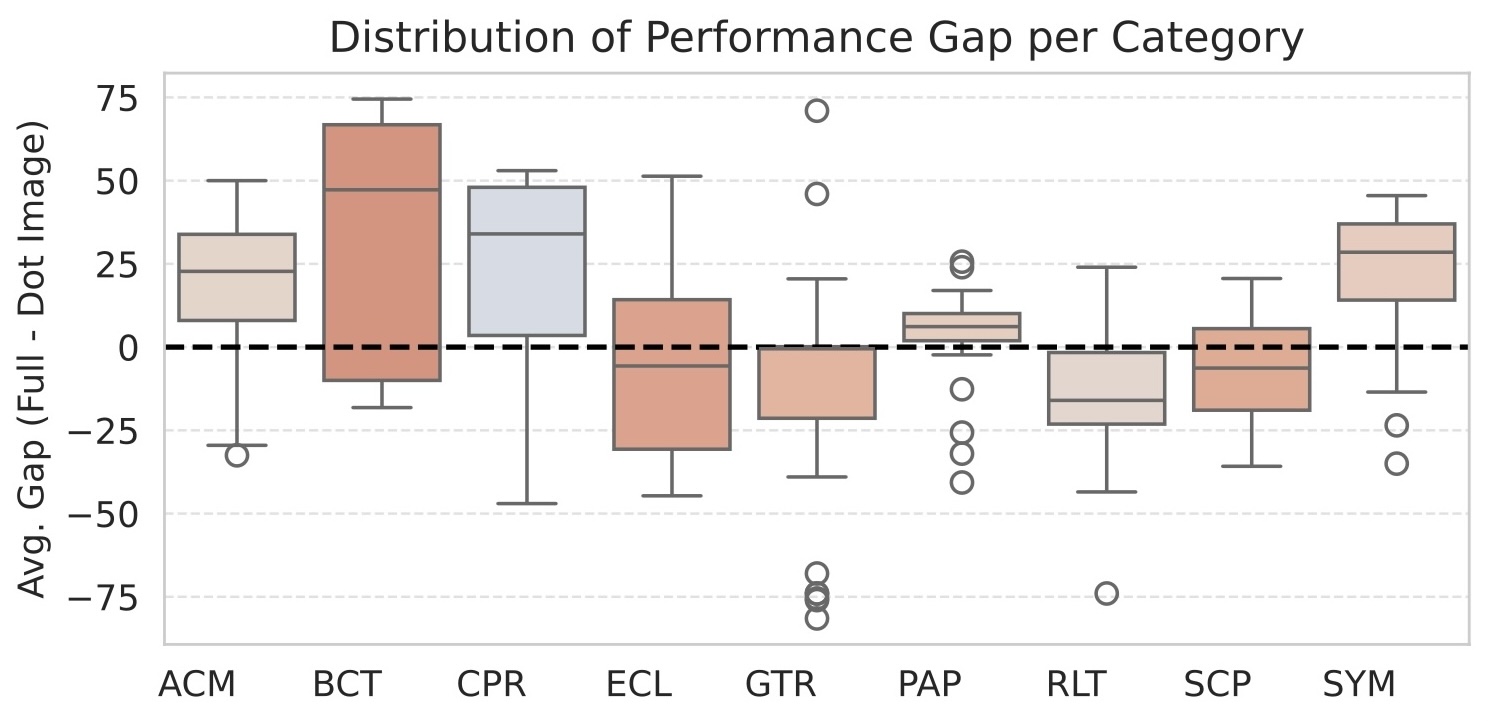}
\caption{Distribution of model-level performance gaps per task category, comparing `text with full image` to `text with dotted image` setups.}
\label{fig:acc_gap_cat_lvl}
\end{figure}

Models evaluated with text and full images consistently outperform both text-only and text with dotted images settings across nearly all task types and categories. For example, Qwen2.5-VL-7B-Instruct reaches 100\% accuracy on several classification and unstable tasks (basic coordinate tasks, symmetry, circle properties) when provided with full image input, compared to much lower scores in the dotted-image setup. It also shows markedly improved performance on numerically intensive tasks such as basic coordinate tasks (79.0\%) and elementary calculations (33.7\%), which are typically challenging across the board.
This strong overall trend is visualized in Figure \ref{fig:acc_gap_cat_lvl}, where most task categories exhibit substantial positive accuracy gaps favoring full images. The largest average gains occur in tasks involving curved shapes and global geometry, such as Area comparison (ACM), Basic Coordinates Tasks (BCT) and Circle Properties (CPR).

By contrast, linear or axis-aligned tasks, such as Parallelism, Perpendicularity, or Geometric Transformations (GTR), show minimal or no improvement between dotted and full visual input. This suggests that dots-only representations already encode sufficient information for solving simpler spatial alignment problems. A closer task-level breakdown further reinforces this distinction: while some tasks yield dramatic gains of +40–100\% when full images are provided, others exhibit near-zero or even negative improvements.

These findings demonstrate that \textbf{high-fidelity visual input is essential for activating geometric reasoning} in current VLMs. While \textbf{sparse or dotted stimuli may suffice for simple linear tasks, they fall short for complex shape recognition and spatial inference}, underscoring the need for more expressive and grounded visual processing in geometric reasoning benchmarks.

\paragraph{Not all VLMs leverage full images equally.}
Figure \ref{fig:acc_gap_cat_lvl} shows the gain from dots to full images. InternVL‑2.5‑1B and Qwen2‑VL‑Instruct improve consistently, whereas InternVL‑3, G‑LLaVA‑13, and URSA gain little or regress, signaling weak visual grounding or instruction‑following most pronounced on numeric items.

This phenomenon may correlate with several factors: poor handling of large or complex images, degraded adherence to structured prompts in multimodal settings, or overfitting to irrelevant visual patterns that misalign with the task objective. These cases highlight a critical limitation -- larger image context can confuse undertrained or improperly aligned VLMs, leading to performance drops.

The takeaway message here is that \textbf{merely accessing visual data is not enough; it's essential to effectively ground and integrate image features to fully capitalize on the advantages offered by complete image input}.

\paragraph{Task sensitivity to modality.} 
Some tasks (e.g., remarkable triangle lines, simple circle properties) show dramatic performance gains when moving from text-only to full visual input (e.g., Qwen2.5: from 0.0\% on text-only task to 99.0\% on full image task on unstable simple circle properties with VL model version). Others (e.g., symmetry) remain relatively stable, indicating that some tasks are more sensitive to modality than others. This uncovers another pattern: \textbf{benchmarking across modality types reveals where geometry is textually recoverable versus inherently visual.}

\paragraph{Instruction-following abilities degradation risk in math-specialized models.} 
Math‑specialized text-only models (e.g., Qwen2.5‑Math‑7B‑Instruct) often ignore structured prompts and misformat outputs, performing worse on classification and unstable tasks than general instruction‑tuned peers. This suggests that domain‑specific fine‑tuning can erode broad instruction adherence by overfitting to rigid mathematical formats.

Additionally, we note the high standard deviations across task categories (see Table \ref{tab:category_agg_results}), which likely reflect the varying difficulty of tasks within each category. Some tasks are straightforward, while others require an understanding of complex geometry and precise calculation of answers, resulting in uneven performance across models. Differences in model capabilities and training objectives also contribute to this variability.

Humans breeze through dotted‑image multiple‑choice items, yet struggle with numeric perimeter/area estimates -- sometimes scoring below InternVL, Phi‑3.5, and LLaVA‑Mini on ECL tasks.

\begin{table*}[]
\centering
\setlength{\tabcolsep}{1mm}
{\fontsize{9pt}{11pt}\selectfont
\begin{tabular}{llrrrrrrrrrrrrrrrrrr}
\toprule
&  & \rotatebox{90}{All$_F$} & \rotatebox{90}{All$_D$} & \rotatebox{90}{C-BCT-CT$_F$} & \rotatebox{90}{C-BCT-CT$_D$} & \rotatebox{90}{U-CPR-CDT$_F$} & \rotatebox{90}{U-CPR-CDT$_D$} & \rotatebox{90}{U-SCP-STT$_F$} & \rotatebox{90}{U-SCP-STT$_D$} & \rotatebox{90}{U-PAP-PLT$_F$} & \rotatebox{90}{U-PAP-PLT$_D$} & \rotatebox{90}{U-PAP-PT$_F$} & \rotatebox{90}{U-PAP-PT$_D$} & \rotatebox{90}{U-PAP-RAT$_F$} & \rotatebox{90}{U-PAP-RAT$_D$} & \rotatebox{90}{C-SYM-SST$_F$} & \rotatebox{90}{C-SYM-SST$_D$} & \rotatebox{90}{U-SYM-ST$_F$} & \rotatebox{90}{U-SYM-ST$_D$} \\
Tran DS &  Config &  &  &  &  &  &  &  &  &  &  &  &  &  &  &  &  &  &  \\
\midrule
All & UF & 97.4 & 58.6 & 98.0 & 50.0 & 100.0 & 52.0 & 100.0 & 48.0 & 94.0 & 77.0 & 93.0 & 61.0 & 100.0 & 68.0 & 100.0 & 67.0 & 94.0 & 46.0 \\
    & FF & 92.6 & 61.4 & 98.0 & 57.0 & 100.0 & 54.0 & 96.0 & 54.0 & 85.0 & 80.0 & 68.0 & 66.0 & 100.0 & 78.0 & 100.0 & 49.0 & 94.0 & 53.0 \\
    & UD & 56.6 & 83.9 & 47.0 & 93.0 & 79.0 & 91.0 & 41.0 & 61.0 & 68.0 & 91.0 & 56.0 & 73.0 & 67.0 & 93.0 & 38.0 & 96.0 & 57.0 & 73.0 \\
    & FD & 49.2 & 72.9 & 47.0 & 92.0 & 51.0 & 66.0 & 48.0 & 59.0 & 59.0 & 84.0 & 43.0 & 61.0 & 49.0 & 84.0 & 38.0 & 74.0 & 59.0 & 63.0 \\
\midrule
Separate & UF & 97.4 & 58.6 & 100.0 & 63.0 & 100.0 & 52.0 & 100.0 & 49.0 & 93.0 & 87.0 & 95.0 & 46.0 & 100.0 & 55.0 & 100.0 & 52.0 & 94.0 & 51.0 \\
    & FF & 92.6 & 61.4 & 99.0 & 48.0 & 100.0 & 52.0 & 100.0 & 51.0 & 88.0 & 83.0 & 70.0 & 58.0 & 100.0 & 55.0 & 100.0 & 52.0 & 91.0 & 49.0 \\
    & UD & 56.6 & 83.9 & 49.0 & 95.0 & 50.0 & 97.0 & 51.0 & 70.0 & 81.0 & 93.0 & 69.0 & 75.0 & 68.0 & 93.0 & 60.0 & 99.0 & 52.0 & 67.0 \\
    & FD & 49.2 & 72.9 & 47.0 & 82.0 & 47.0 & 67.0 & 49.0 & 61.0 & 71.0 & 87.0 & 58.0 & 68.0 & 77.0 & 90.0 & 41.0 & 89.0 & 60.0 & 65.0 \\
\bottomrule
\end{tabular}
}
\caption{Linear‑probe accuracy (binary) for models trained jointly (top 4 rows) or per task (bottom 4). Configs: UF/FF = unfrozen/frozen encoder on full images; UD/FD = unfrozen/frozen on dot images. Evaluation tags: BC = basic‑coordinate, CP = circle‑properties, PP = parallelism‑perpendicularity, Sym = symmetry; subscript F/D marks full vs. dot test images.}
\label{tab:ve_all}
\end{table*}

\subsection{Vision Encoders Linear Probing}

To measure how much of each geometry task is already linearly separable in contemporary vision embeddings \cite{alain2017understanding}, we carry out 2 stages of linear probing.

\paragraph{Estimating task solvability in standalone vision features.}
We extract image embeddings with the OpenAI CLIP‑ViT-B/32 encoder \cite{radford2021} and train a linear classifier on them. For every one of eight binary geometry tasks -- collinearity, diameter, semicircle‑triangle, parallelism, perpendicular lines, right angles, shape symmetry, and symmetry points -- we generate 10K training and 10K test images. In the multi‑task regime this amounts to 80K training samples.

We probe each VLM's vision backbone and CLIP to assess (i) tasks' linear separability and (ii) geometric improvements from vision-language pre-training. Four setups are tested: frozen/fine-tuned encoders with full/dot-only images, using linear heads pooling all patch tokens. Training includes task-specific probes (10K samples each), multi-task probes (80K images), and cross-task transfer tests.

\subsection{Probing Results Analysis}

\paragraph{Full vs.\ dotted diagrams.}
A fine-tuned ViT-B/32 linear probe achieves 97\% accuracy on full images but drops to 58\% on dot-only tests, showing CLIP's reliance on global shape cues. Retraining on dots reverses this pattern: 85\% on dots, 57\% on full images.

\paragraph{Effect of freezing.} Freezing the backbone costs roughly five points on full images (FF = 92\%) and yields the weakest dot performance (FD $\sim$ 73\%), indicating that a small amount of adaptation is important for geometry.

\paragraph{Cross‑task transfer (Table \ref{tab:ve-transfer}).}
Parallel‑line probes push right‑angle accuracy to 86\%, and symmetry probes lift circle‑symmetry to 81\%, confirming family‑level transfer. Still, VLMs lag on NoReGeo, implying language training overlooks these geometric cues.

\begin{table}[h!]
    \centering
    {\fontsize{9pt}{11pt}\selectfont
     \begin{tabular}{lll}
        \toprule
        Train Dataset & Eval Dataset & Configs with Accuracy \\
        \midrule
        U-SYM-ST & C-SYM-SST & UD (81.0) \\
        U-SYM-ST & U-CPR-CDT & UD (74.0) \\
        C-SYM-SST & U-CPR-CDT & FF (79.0) \\
        C-SYM-SST & C-BCT-CT & FD (86.0) \\
        U-PAP-PLT & U-PAP-PT & \parbox{3.5cm}{UD (75.0), UD (74.0), UF (71.0)} \\
        U-PAP-PLT & C-BCT-CT & FD( 90.0), FF (78.0) \\
        U-PAP-PT & C-SYM-SST & FF (72.0) \\
        U-PAP-PT & U-PAP-PLT & \parbox{3.5cm}{UD (90.0), UD (86.0), FD (85.0)} \\
        U-PAP-PT & U-PAP-RAT & \parbox{3.5cm}{UD (80.0), FD (78.0), UF (71.0)} \\
        U-PAP-PT & U-CPR-CDT & \parbox{3.5cm}{UD (83.0), FF (76.0), UF (75.0)} \\
        U-SCP-STT & C-SYM-SST & FF (90.0) \\
        U-CPR-CDT & U-SCP-STT & UF (80.0), FF (77.0) \\
        C-BCT-CT & U-SCP-STT & UF (95.0) \\
        C-BCT-CT & U-CPR-CDT & UF (100.0), FF (73.0) \\
        \bottomrule
    \end{tabular}
    }
    \caption{Linear‐probing training transfer.}
    \label{tab:ve-transfer}
\end{table}

\section{Conclusion}

We introduced NoReGeo, a cross-modal benchmark of elementary geometry problems designed to assess whether LLMs and VLMs can answer spatial questions \emph{without} relying on explicit reasoning. Across more than 45 state-of-the-art models, we find that most struggle with tasks that humans solve through immediate geometric intuition, often producing unnecessary chain-of-thought explanations even when instructed otherwise. In contrast, a frozen vision encoder paired with a simple linear probe performs almost perfectly, indicating that the essential geometric cues are already embedded in visual representations.

These findings position NoReGeo as a practical tool for probing latent geometric competence in modern foundation models and for selecting models when fast, geometry-aware inference is required. Looking ahead, we plan to explore how fine-tuning and representation alignment influence generalization across geometric concepts and whether models can be encouraged to develop more robust, human-like geometric intuition.

\appendix

\section{Acknowledgments}
Innopolis University authors were supported by the Research Center of the Artificial Intelligence Institute at Innopolis University. Financial support was provided by the Ministry of Economic Development of the Russian Federation (No. 25-139-66879-1-0003).

\bibliography{aaai2026}

\newpage 

\appendix

\section{Appendix A. Evaluation details and models performance}
\label{sec:suppl:res-model}

\subsection{Evaluation protocol.} Table \ref{tab:system_prompt} lists the system prompts used for every task. All inferences were run at temperature 0 (no sampling) to ensure consistent results.

\begin{table}[ht!]
\centering
\begin{tabular}{p{0.95\linewidth}}
\toprule
\textbf{System Prompt} \\
\midrule
{\raggedright\ttfamily
You are a highly capable AI assistant with expertise in geometry. You can accurately analyze geometric figures, solve problems.
} \\
\toprule
\textbf{Structuring Prompt Variations} \\
\midrule
\textbf{[Point answer type]:} {\raggedright\ttfamily Provide your answer as JSON with keys: 'x' and 'y' for point coordinates. Return only that object.}	\\
\midrule
\textbf{[Number answer type]:} {\raggedright\ttfamily Provide your answer as JSON: \{'answer': <value>\}, where <value> is a floating point or integer number. Return only that object.}	\\
\midrule
\textbf{[Multiple-choice answer type]:} {\raggedright\ttfamily Provide your answer as JSON: \{'answer': <value>\}, where <value> is from the options: ABC, DEF, equal. Return only that object.} \\
\midrule
\textbf{[Multiple-choice answer type]:} {\raggedright\ttfamily Provide your answer as JSON: \{'answer': <value>\}, where <value> is from the options: ABCD, EFGH, equal. Return only that object.} \\
\midrule
\textbf{[Multiple-choice answer type]:} {\raggedright\ttfamily Provide your answer as JSON: \{'answer': <value>\}, where <value> is from the options: median, altitude, bisector. Return only that object.} \\
\midrule
\textbf{[Multiple-choice answer type]:} {\raggedright\ttfamily Provide your answer as JSON: \{'answer': <value>\}, where <value> is from the options: yes, no. Return only that object.} \\
 \\
\bottomrule
\end{tabular}
\caption{System and answer-structuring prompts used during inference.}
\label{tab:system_prompt}
\end{table}

\subsection{Detailed model evaluation}

\begin{figure*}[ht!]
  \centering
  \begin{subfigure}[t]{0.9\textwidth}
    \centering
    \includegraphics[width=\textwidth]{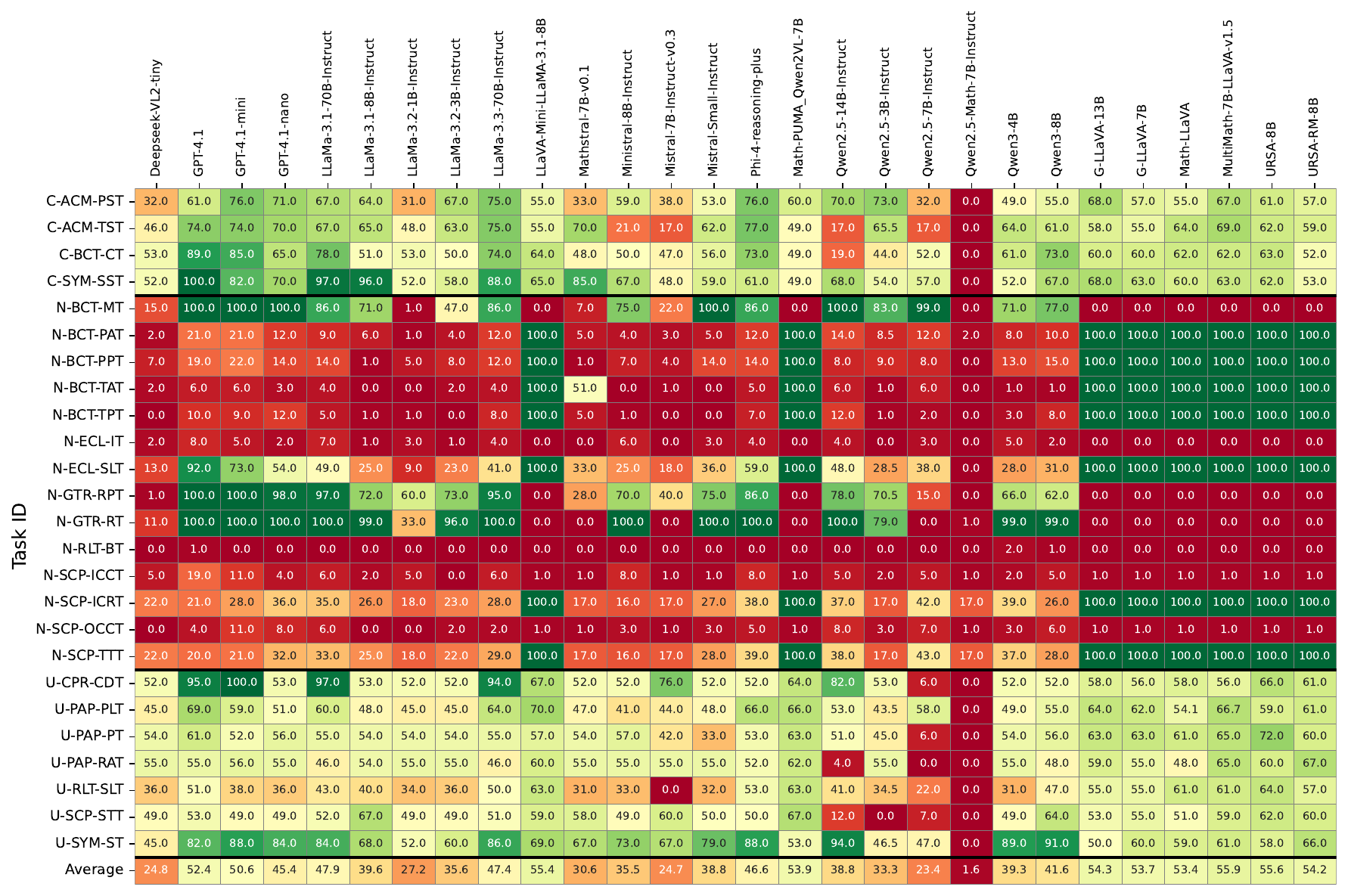}
    \caption{
      Accuracy heatmap for the NoReGeo benchmark with text-only input.
      Rows correspond to individual tasks, columns to LLMs. 
      Cell color intensity reflects model performance: greener indicates higher accuracy, redder indicates lower performance.
    }
  \end{subfigure}
  \hfill
  \begin{subfigure}[t]{0.9\textwidth}
    \centering
    \includegraphics[width=\textwidth]{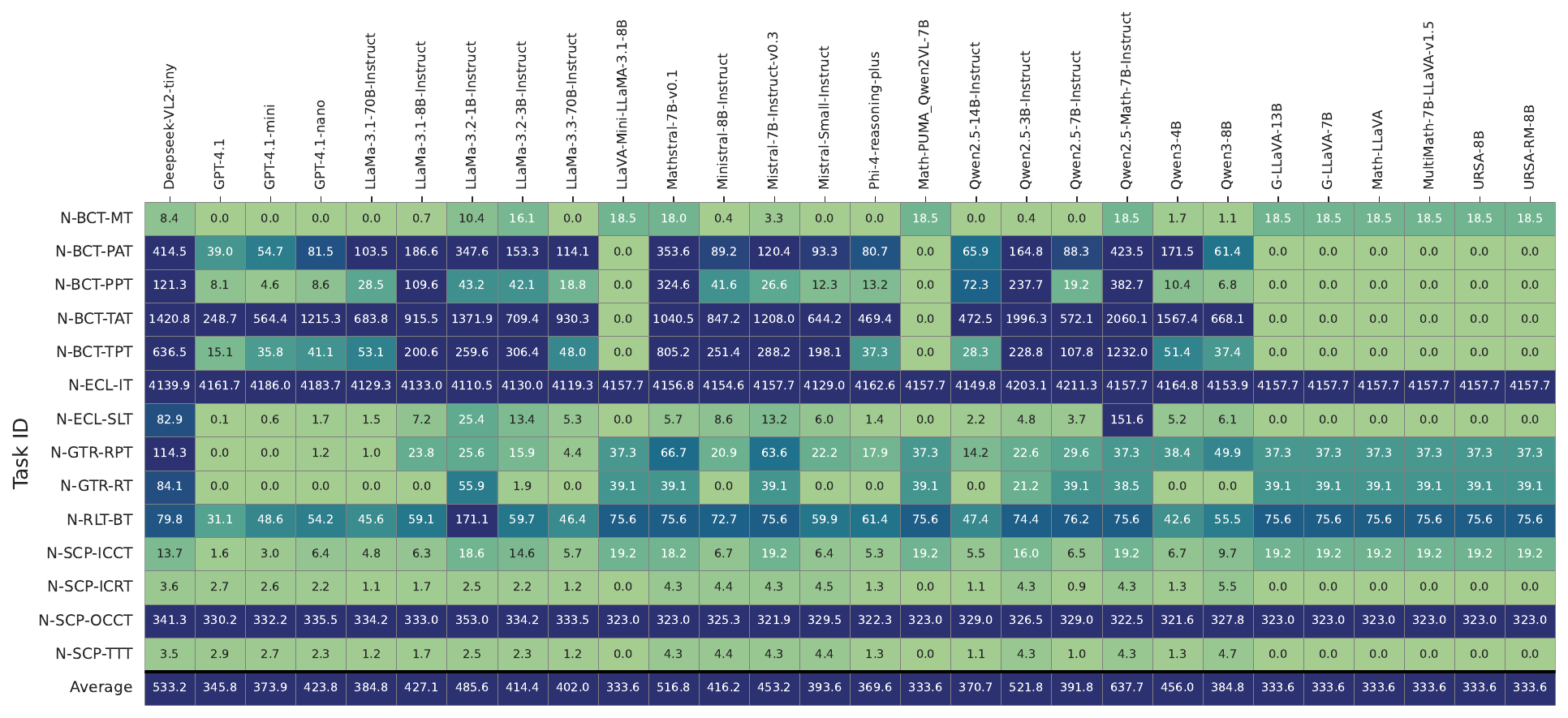}
    \caption{
      Mean Squared Error (MSE) heatmap on numeric tasks under the text-only NoReGeo setup.
      Each row represents a numeric reasoning task, and each column an LLM.
      Greener cells indicate lower regression error, while more blue cells highlight higher error.
    }
  \end{subfigure}
  \caption{
    Detailed performance evaluation of large language models (LLMs) on the NoReGeo benchmark using only text inputs.
    Subfigure (a) shows task-level classification accuracy across all task types, while subfigure (b) focuses on regression error (MSE) for numeric tasks.
  }
  \label{fig:detailed_noregeo_results_text}
\end{figure*}

In this section, we present detailed heatmaps of model performance across all NoReGeo task categories, types and evaluation setups. 

\begin{figure*}[h]
  \centering
  \begin{subfigure}[t]{0.9\textwidth}
    \centering
    \includegraphics[width=\textwidth]{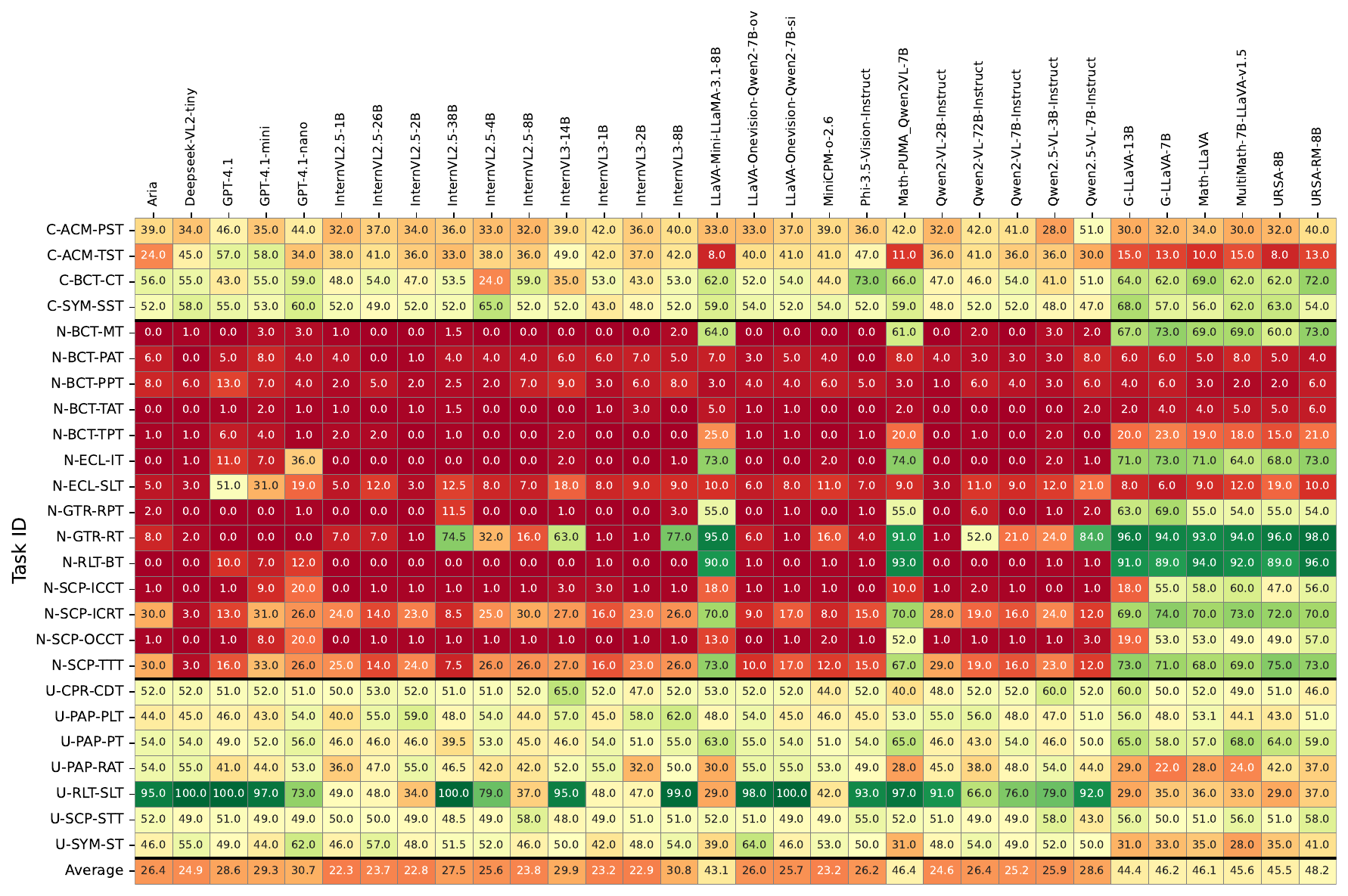}
    \caption{
      Accuracy heatmap for the NoReGeo benchmark with `text + dotted images` input.
      Rows correspond to individual tasks, columns to LLMs. 
      Cell color intensity reflects model performance: greener indicates higher accuracy, redder indicates lower performance.
    }
  \end{subfigure}
  \hfill
  \begin{subfigure}[t]{0.9\textwidth}
    \centering
    \includegraphics[width=\textwidth]{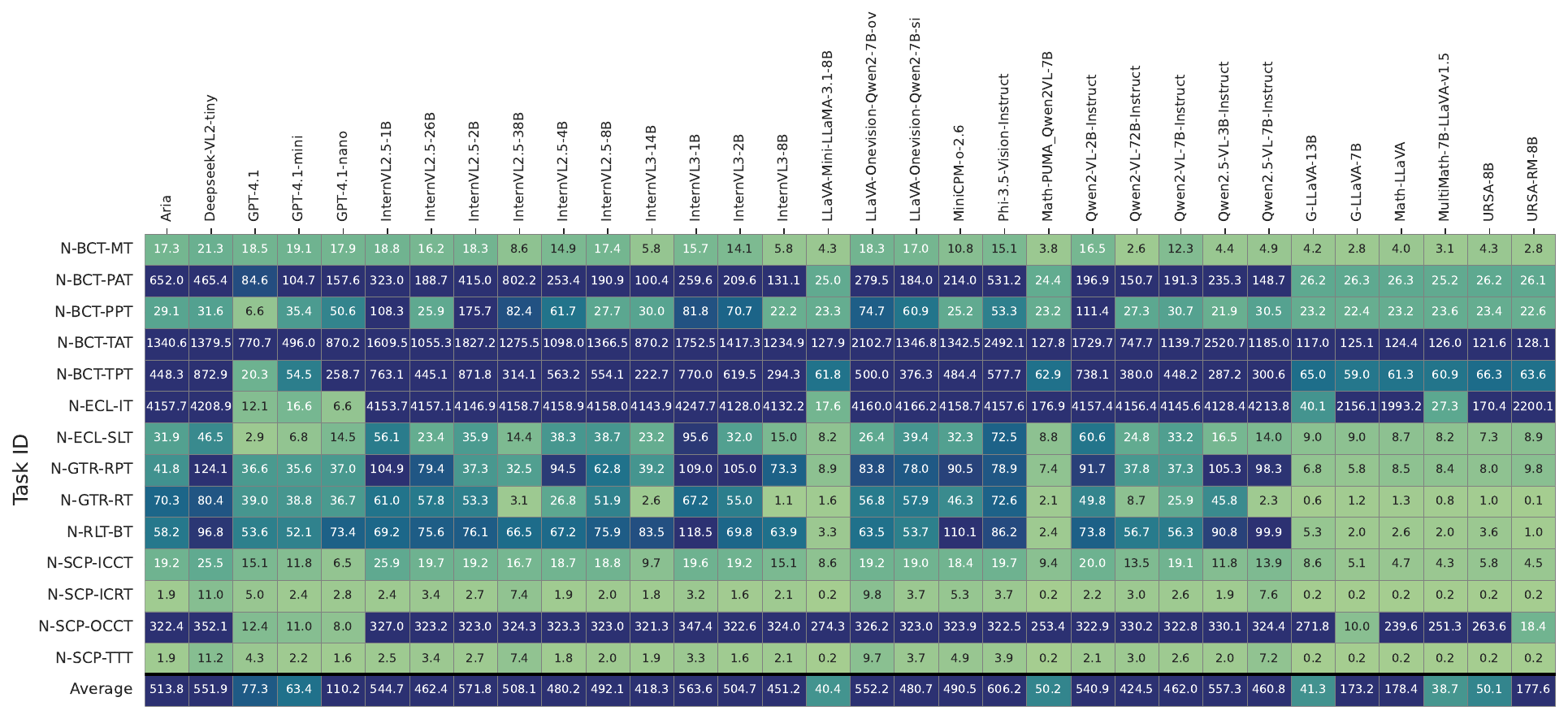}
    \caption{
      Mean Squared Error (MSE) heatmap on numeric tasks under the `text + dotted images` NoReGeo setup.
      Each row represents a numeric reasoning task, and each column an LLM.
      Greener cells indicate lower regression error, while more blue cells highlight higher error.
    }
  \end{subfigure}
  \caption{
    Detailed performance evaluation of large language models (LLMs) on the NoReGeo benchmark using `text + dotted images` inputs.
    Subfigure (a) shows task-level classification accuracy across all task types, while subfigure (b) focuses on regression error (MSE) for numeric tasks.
  }
  \label{fig:detailed_noregeo_results_dot}
\end{figure*}

\begin{figure*}[h]
  \centering
  \begin{subfigure}[t]{0.9\textwidth}
    \centering
    \includegraphics[width=\textwidth]{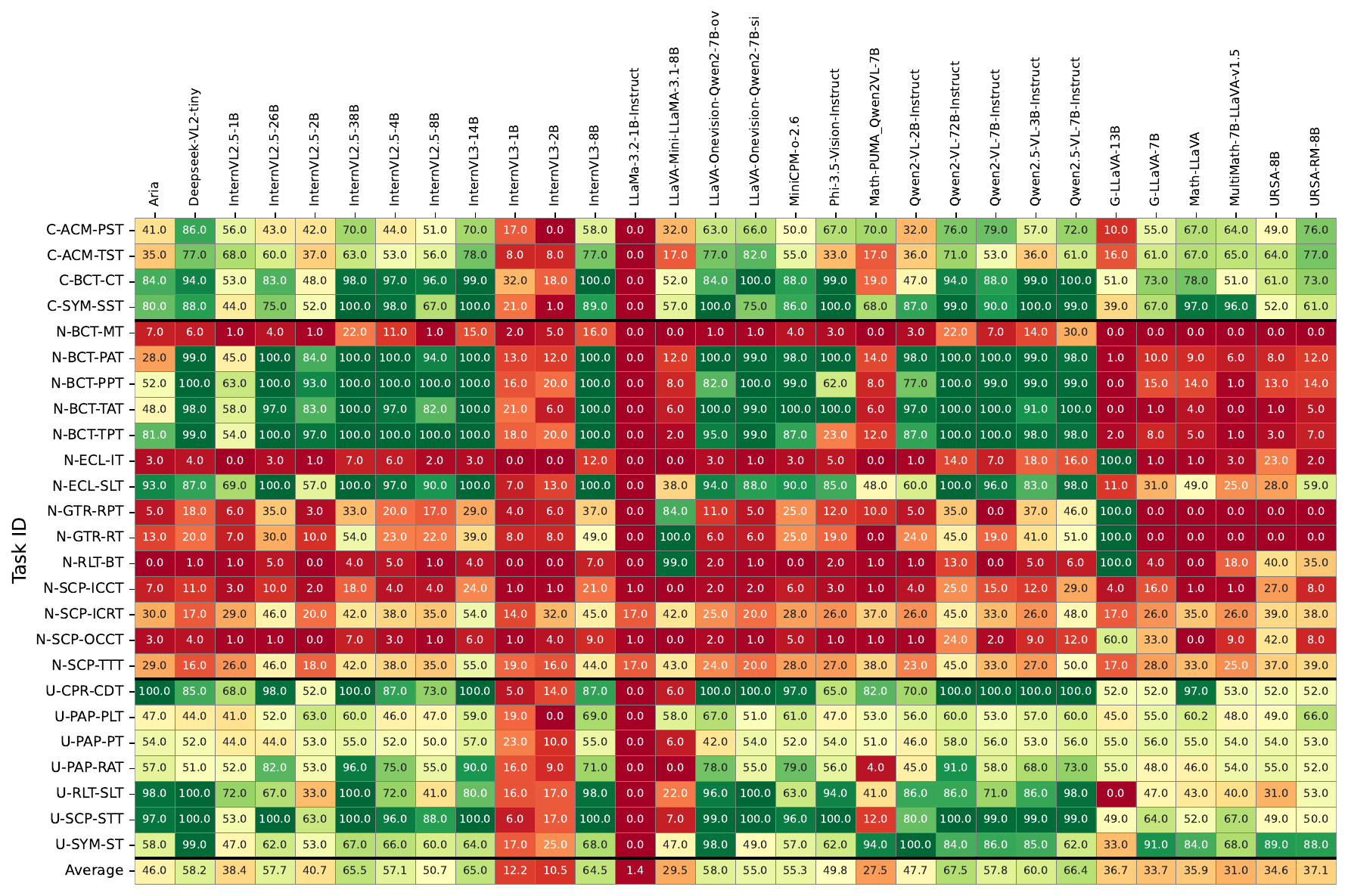}
    \caption{
      Accuracy heatmap for the NoReGeo benchmark with `text + full images` input.
      Rows correspond to individual tasks, columns to LLMs. 
      Cell color intensity reflects model performance: greener indicates higher accuracy, redder indicates lower performance.
    }
  \end{subfigure}
  \hfill
  \begin{subfigure}[t]{0.9\textwidth}
    \centering
    \includegraphics[width=\textwidth]{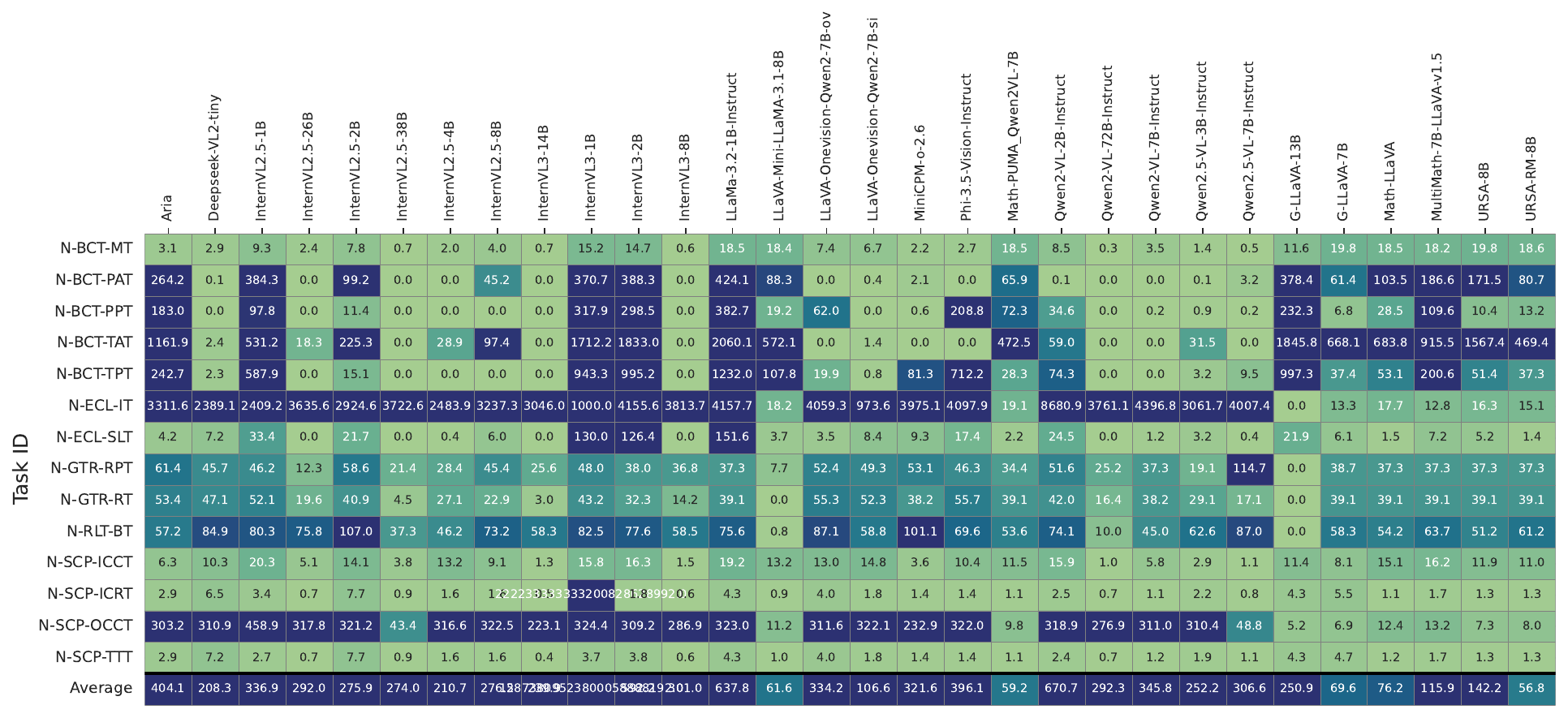}
    \caption{
      Mean Squared Error (MSE) heatmap on numeric tasks under the `text + full images` NoReGeo setup.
      Each row represents a numeric reasoning task, and each column an LLM.
      Greener cells indicate lower regression error, while more blue cells highlight higher error.
    }
  \end{subfigure}
  \caption{
    Detailed performance evaluation of large language models (LLMs) on the NoReGeo benchmark using `text + full images` inputs.
    Subfigure (a) shows task-level classification accuracy across all task types, while subfigure (b) focuses on regression error (MSE) for numeric tasks.
  }
  \label{fig:detailed_noregeo_results_full}
\end{figure*}

In addition to the accuracy-based evaluations shown in Figure \ref{fig:detailed_noregeo_results_text} (a), Figure \ref{fig:detailed_noregeo_results_dot} (a), Figure \ref{fig:detailed_noregeo_results_full} (a), we include regression-based heatmaps using Mean Squared Error (MSE) to assess the numerical precision of model outputs in tasks requiring quantitative responses: Figure \ref{fig:detailed_noregeo_results_text} (b), Figure \ref{fig:detailed_noregeo_results_dot} (b) and Figure \ref{fig:detailed_noregeo_results_full} (b). 

These plots reveal complementary insights by highlighting cases where models may produce approximately correct values despite low classification accuracy, or vice versa. The MSE metric is particularly informative for tasks with numerical and coordinates answer types, where small deviations from the correct answer carry semantic significance.

\begin{figure*}[] 
\centering
\includegraphics[width=0.9\textwidth]{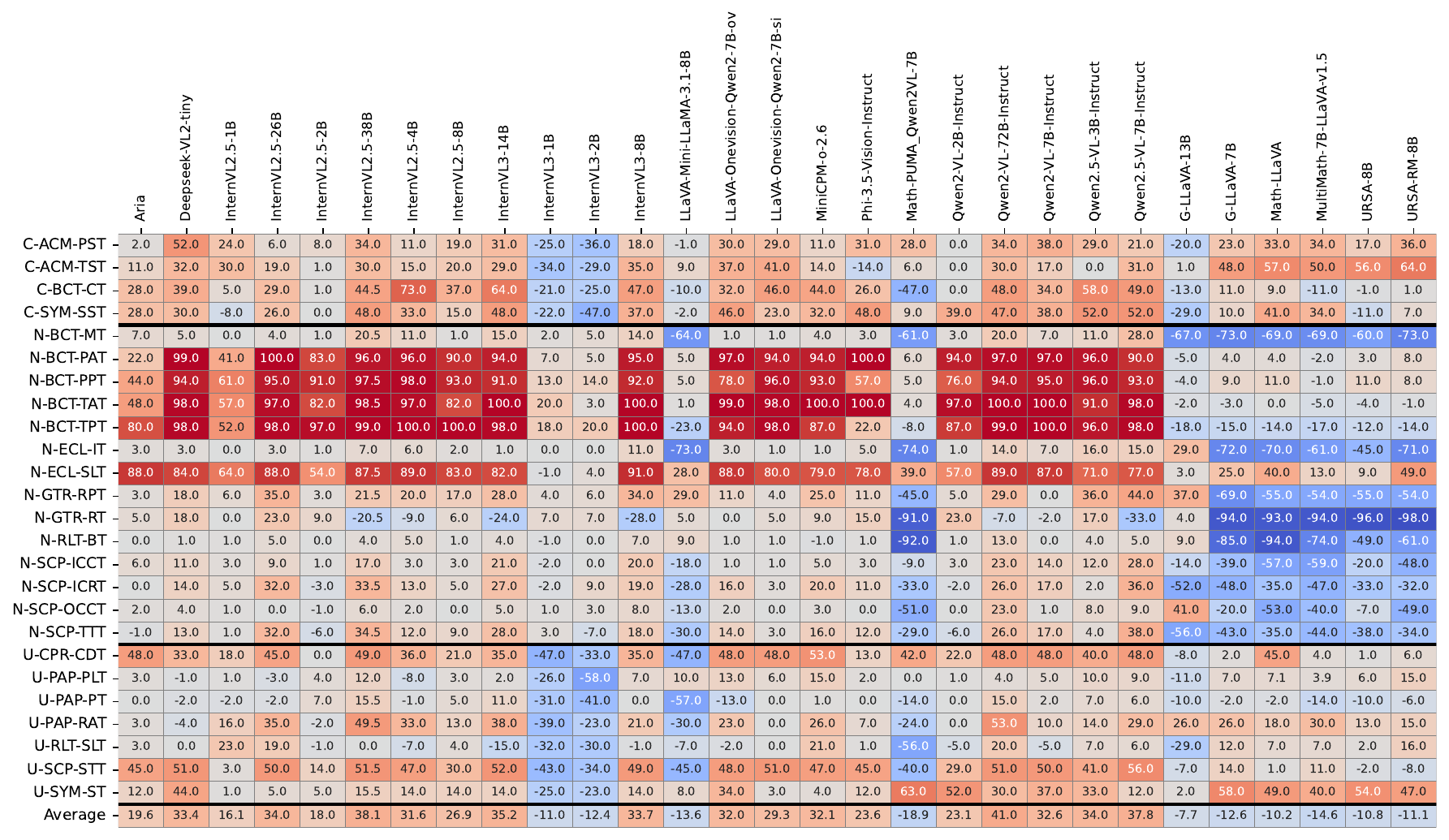}
\caption{Performance gap between full-image and dot-image setups for VLMs.
This heatmap shows the difference in Accuracy rate (\%) for each task when evaluated with full images vs. dot images, computed as: Accuracy(full) − Accuracy(dot). Blue cells indicate improved performance with full images, red cells indicate a drop. The final row shows the average difference between setups across tasks for each model.}
\label{fig:acc_gap_heatmap}
\end{figure*}

\begin{figure*}[] 
\centering
\includegraphics[width=0.8\textwidth]{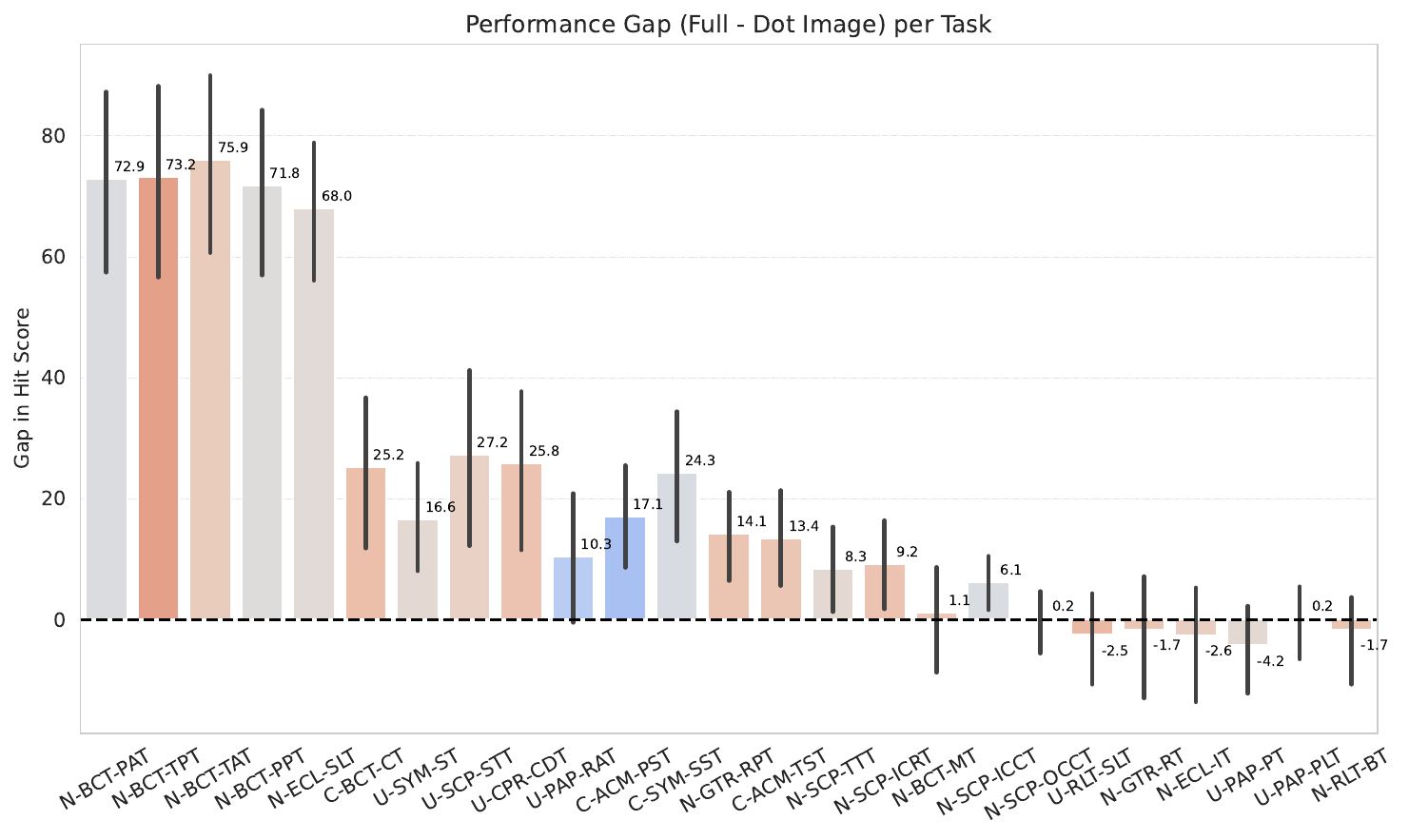}
\caption{Task-level performance difference between the `text + full image` and `text + dot image` evaluation setups for VLMs.}
\label{fig:acc_gap_task_lvl}
\end{figure*}

As discussed in Main results section, models evaluated with text + full images generally outperform those using text + dotted images across most task types and categories. However, a more detailed analysis is crucial, as the performance gap varies significantly across conceptually different tasks. To capture this variation, we provide a task-level summary of performance differences between the two visual setups in Figure~\ref{fig:acc_gap_task_lvl}, along with a fine-grained heatmap of model-by-task performance gaps in Figure~\ref{fig:acc_gap_heatmap}.

\section{Appendix B. Visual Encoders Experimental Details}\label{app:training-details}

\subsection{Architecture and Implementation}

Our experimental framework employs the OpenAI CLIP-ViT-base-patch32 model as the foundation visual encoder, utilizing its pre-trained representations learned from large-scale image-text pairs. The architecture processes input images at 224×224 resolution with 32×32 patch decomposition, resulting in a sequence of 49 patch tokens plus one classification token.

For feature extraction, we implement a linear pooling strategy that aggregates information across all spatial locations. Given the transformer's output embeddings of dimensionality $[\text{batch\_size}, 50, 768]$, we apply global flattening to obtain $[\text{batch\_size}, 38400]$ and subsequently project through a learnable linear transformation to the original embedding dimension $[\text{batch\_size}, 768]$. This approach allows the model to learn optimal spatial attention weights rather than relying on fixed pooling strategies.

The classification head consists of a single linear layer mapping from the 768-dimensional pooled representation to binary output logits. We systematically evaluate two training regimes: frozen backbone configuration where only the linear layers are optimized, and fine-tuned configuration allowing end-to-end parameter updates throughout the visual encoder.

Our implementation leverages PyTorch framework with Hugging Face Transformers library for model instantiation and management. Training is conducted on single GPU with automatic mixed precision to optimize memory utilization and computational efficiency. We employ comprehensive logging through TensorBoard to monitor training dynamics, convergence behavior, and per-task performance metrics throughout the optimization process.

\subsection{Training Configuration}

All experiments follow a consistent training protocol designed for rapid convergence on geometric perception tasks. We utilize the AdamW optimizer with a conservative learning rate of $2 \times 10^{-6}$ and L2 regularization coefficient of 0.01 to prevent overfitting. The training spans 2 epochs with batch size 32, sufficient for convergence given the relatively simple linear classification objective. All experiments were conducted on single A100 (80GB) GPU.

Early stopping mechanism monitors validation F1-score, prioritizing model generalization over training loss minimization. Cross-entropy loss serves as the optimization objective for all binary classification tasks. Complete hyperparameter settings are summarized in Table~\ref{tab:hyperparams}.

The complete training and evaluation results across the full experimental matrix are presented in Figure~\ref{fig:VE_probing}, providing comprehensive accuracy scores for each train-evaluation task pair under all model configurations.

\begin{table}[h]
\centering
\begin{tabular}{ll}
\toprule
\textbf{Parameter} & \textbf{Value} \\
\midrule
Batch size & 32 \\
Epochs & 2 \\
Learning rate & $2 \times 10^{-6}$ \\
Weight decay & 0.01 \\
Optimizer & AdamW \\
Loss function & Cross-entropy \\
Early stopping metric & Validation F1-score \\
Mixed precision & Enabled \\
\bottomrule
\end{tabular}
\caption{Training hyperparameters for visual encoder probing experiments}
\label{tab:hyperparams}
\end{table}

\subsection{Complete Results}

\begin{figure*}[h] 

\includegraphics[width=1\textwidth]{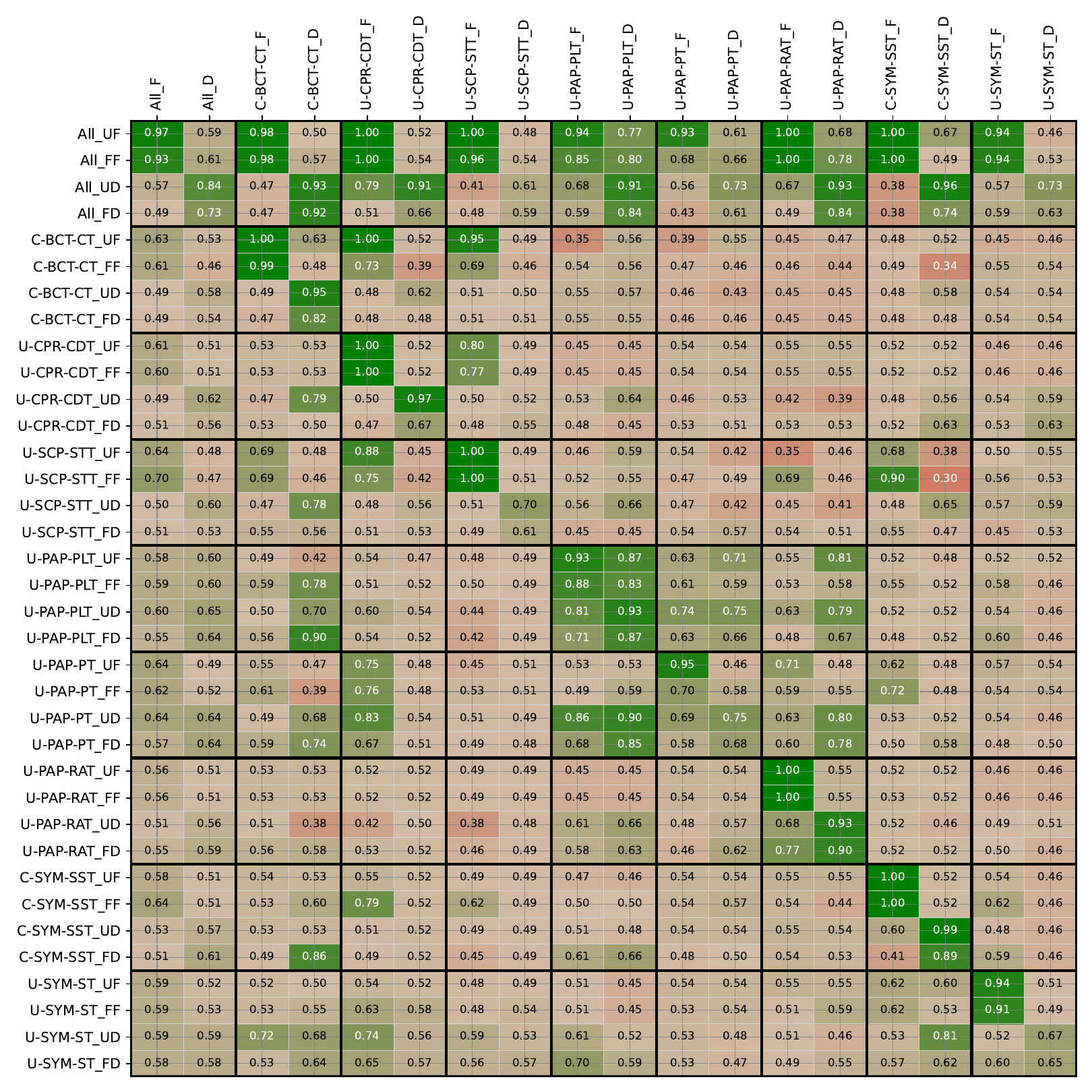}
\caption{Appendix B Table of results for all experiments on ViT model.}
\label{fig:VE_probing}
\end{figure*}

The complete training and evaluation results across the full experimental matrix are presented in Figure~\ref{fig:VE_probing}, providing comprehensive accuracy scores for each train-evaluation task pair under all model configurations.

\section{Appendix C. Human Evaluation Protocol}

Tasks from the NoReGeo benchmark were preprocessed for the Toloka platform. We used the dotted version of the benchmark split; ground truth answers were excluded and stored separately. Training, examination, and control tasks were created for the annotators. The tasks' creation protocol is given below.

\begin{itemize}
    \item Each page contained 10 tasks, and annotators had 10 minutes to complete a page.
    \item We applied majority voting with an overlap of 10: for each task, votes for all options were tallied.
    \item We computed the same evaluation metrics on the aggregated annotation table as those used for model evaluation.
    \item The average annotator age was 39 years, and the compensation averaged \$1 per task page.
\end{itemize}

\end{document}